\documentclass[conference]{IEEEtran}
\IEEEoverridecommandlockouts
\usepackage{cite}
\usepackage{amsmath,amssymb,amsfonts}
\usepackage{algorithm}
\usepackage{algpseudocode}
\usepackage{graphicx}
\usepackage{textcomp}
\usepackage{xcolor}
\def\BibTeX{{\rm B\kern-.05em{\sc i\kern-.025em b}\kern-.08em
    T\kern-.1667em\lower.7ex\hbox{E}\kern-.125emX}}
\usepackage{tikz}

\newcommand\copyrighttext{%
  \footnotesize \textcopyright 2024 IEEE.   Personal use of this material is permitted.  Permission from IEEE must be obtained for all other uses, in any current or future media, including reprinting/republishing this material for advertising or promotional purposes, creating new collective works, for resale or redistribution to servers or lists, or reuse of any copyrighted component of this work in other works.}
\newcommand\copyrightnotice{%
\begin{tikzpicture}[remember picture,overlay]
\node[anchor=south,yshift=10pt] at (current page.south) 
  {\fbox{\parbox{\dimexpr\textwidth-\fboxsep-\fboxrule\relax}{\copyrighttext}}};
\end{tikzpicture}%
}

\def\BibTeX{{\rm B\kern-.05em{\sc i\kern-.025em b}\kern-.08em
    T\kern-.1667em\lower.7ex\hbox{E}\kern-.125emX}}
\begin{document}

\title{Fast, Private, and Protected: Safeguarding Data Privacy and Defending Against Model Poisoning Attacks in Federated Learning \thanks{This is the author's accepted manuscript of the article. The version of record is available at IEEE Xplore: NRG Assumpcao, "Fast, Private, and Protected: Safeguarding Data Privacy and Defending Against Model Poisoning Attacks in Federated Learning" in 2024 IEEE Symposium on Computers and Communications (ISCC), DOI: 10.1109/ISCC61673.2024.10733713.}}

\author{\IEEEauthorblockN{Nicolas Riccieri Gardin Assumpcao}
\IEEEauthorblockA{\textit{Institute of Computing} \\
\textit{State University of Campinas}\\
Campinas, Brazil \\
n121245@dac.unicamp.br}
\and
\IEEEauthorblockN{Leandro Villas}
\IEEEauthorblockA{\textit{Institute of Computing} \\
\textit{State University of Campinas}\\
Campinas, Brazil \\
lvillas@ic.unicamp.br}
}

\maketitle
\copyrightnotice
\begin{abstract}
Federated Learning (FL) is a distributed training paradigm wherein participants collaborate to build a global model while ensuring the privacy of the involved data, which remains stored on participant devices. However, proposals aiming to ensure such privacy also make it challenging to protect against potential attackers seeking to compromise the training outcome. In this context, we present Fast, Private, and Protected (FPP), a novel approach that aims to safeguard federated training while enabling secure aggregation to preserve data privacy. This is accomplished by evaluating rounds using participants' assessments and enabling training recovery after an attack. FPP also employs a reputation-based mechanism to mitigate the participation of attackers. We created a dockerized environment to validate the performance of FPP compared to other approaches in the literature (FedAvg, Power-of-Choice, and aggregation via Trimmed Mean and Median). Our experiments demonstrate that FPP achieves a rapid convergence rate and can converge even in the presence of malicious participants performing model poisoning attacks.
\end{abstract}


\section{Introduction} \label{sec:introduction}

Federated Learning (FL) is a decentralized machine learning paradigm conducted across distributed devices, which only shares updated parameters during training. This technique eliminates data centralization, as the data remains stored on participant devices.

FL involves a group of devices (referred to as clients) with their respective datasets performing distributed training. In each interaction (known as round), clients receive the global model, execute several epochs of local training using their private datasets, and then return the trained models to be aggregated into a new global model. This allows training to occur without the need for data to leave the devices.

This characteristic provides an alternative for model training under new data privacy regulations, such as the General Data Protection Regulation (GDPR) in the European Union, the California Consumer Privacy Act in the state of California, and the General Data Protection Law (LGPD) in Brazil.

Nevertheless, the utilization of private data and the distributed aspect of FL create significant challenges: the diversity of the clients' data, how to guarantee the privacy of this data, and how to protect the training against attacks conducted by malicious clients.

Since the clients train in parallel using only the data regarding their context, this may cause the datasets to be non-independent and identically distributed (non-iid). This data heterogeneity negatively impacts the convergence speed of the federated model\cite{cho2020client}.

The guarantee of data privacy is also a significant challenge. Research such as \cite{zhu2019deep} has shown how it is possible to reconstruct the dataset using only the gradients shared by the clients during training.

Another challenge that emerges from the distribution of FL training is the possibility of attacks aimed at sabotaging the final result, such as the model poisoning attack, which compromises aggregation results, wasting the efforts of honest participants in creating an efficient and useful global model \cite{lyu2020threats} \cite{nair2023robust}. Techniques to protect training against this type of attack may include auditing the content of gradients to detect signs of tampering\cite{kang2019incentive}.

Protecting data privacy against leakage and training against attacks simultaneously is complex because both problems seem to require opposing approaches. On one hand, gradients are concealed to protect data, while on the other hand, techniques are developed to analyze gradients to detect attack signals.

In this context, we introduce Fast Private and Protected (FPP), a new approach for FL that improves the client selection strategy to enhance training efficiency even in scenarios with non-iid data. We also propose a new evaluation method for scoring training compatible with Secure Aggregation\cite{bonawitz2017practical}. The FPP also enables the recovery of a prior checkpoint of the global model in cases of severe attacks, preserving the training evolution.

The next section presents related works, highlighting their contributions and limitations. Section 3 introduces the FPP and describes how it mitigates problems associated with non-iid data, model poisoning attacks, and data privacy. Section 4 describes the experiment setup and the environment developed to validate FPP. Section 5 provides the experiment results and analysis, and Section 6 concludes this work by presenting future opportunities for research.

\section{Related Work} \label{sec:correlatos}

There are various challenges in the FL context due to its distributed nature and the involvement of sensitive, sometimes non-iid, data. This section presents some solutions from the literature aimed at addressing these challenges.

FedAvg \cite{brendan2016communication} proposes running several epochs of training on clients to generate new local models for aggregation into a new global model. This strategy accelerates FL training but may not be suitable for cases where client data is non-iid.

Power-of-Choice (PoC) \cite{cho2020client} introduces an approach to client selection that is more sophisticated, allowing for the reduction of the number of clients training in each round and accelerating convergence. PoC achieves this by focusing training on clients with the highest loss values, and prioritizing those with worse metrics, as they have more room for improvement in collaboration with the global model. PoC is particularly effective when applied to non-iid data, as the heterogeneity among clients increases the variation in learning potential. However, PoC does not offer protection against malicious participants.

To protect user data privacy, secure aggregation \cite{bonawitz2017practical} employs Homomorphic Encryption to mask gradients received from each client while preserving their sum, ensuring aggregation consistency. This is achieved through a protocol where controlled noises are added to gradients in a manner that cancels them out during summation.

While secure aggregation prevents the disclosure of shared gradients, it also hinders gradient auditing to verify their legitimacy. Therefore, it is incompatible with methods like \cite{kang2019incentive}, which analyze gradients for unusual distributions to detect attacks.

Certain works such as \cite{kang2019incentive}, \cite{wang2020optimizing}, and \cite{zhang2021client} utilize a public dataset to assess the quality of an aggregated model, detecting model poisoning attacks by observing metric deterioration post-evaluation with the public dataset. However, finding a public dataset that adequately represents real data can be challenging within the FL context.

Other proposals to protect against model poisoning attacks suggest alternatives to aggregation metrics. \cite{yin2018byzantine} uses the median or a trimmed mean for gradient aggregation, \cite{pillutla2019robust} calculates the geometric median to find the center of mass of the gradients, and \cite{xie2018generalized} aggregates only gradients near the median. These methods exclude outliers during aggregation. However, they are impractical to use with secure aggregation, which preserves the sum and average but makes calculating the median or trimmed mean unfeasible.

\cite{zhang2023reputation} and \cite{jiang2024research} propose a reputation mechanism to evaluate the contribution of each client participant in each round to remove clients that degrade the training. However, assessing each client's contribution is also incompatible with secure aggregation since it is impossible to validate each client's contribution individually.

These related works exemplify how privacy protection, training with non-iid data, and the potential presence of attackers are only partially addressed by the literature, as they typically focus on one of these problems. Particularly, simultaneously addressing data privacy and attack protection appears to involve a trade-off between concealing or analyzing gradients.

To mitigate this limitation, we introduce the FPP. This new FL framework aims to retain the benefits of efficient client selection for accelerating training convergence while simultaneously addressing protection against model poisoning attacks and gradient leakage. It achieves this by enabling the use of secure aggregation to conceal individual client gradients.

\section{Introducing FPP} \label{sec:proposta}

The FPP was developed to simultaneously mitigate convergence issues using non-iid data, ensure the privacy of clients' datasets, and prevent model poisoning attacks. This is achieved through the following characteristics:

\begin{itemize}
    \item Focusing training on clients with the highest loss values, where there is the greatest potential for improvement.   
    \item Evaluating the global model using real datasets from clients.
    \item Implementing recovery of a previous checkpoint of the global model after an attack.
    \item Utilizing a reputation value to mitigate the participation of malicious clients.
    \item The gradient aggregation, the model performance evaluation, and the reputation estimation of each client don't use the individual gradients, allowing the utilization of Secure Aggregation \cite{bonawitz2017practical} to protect the data privacy. 
\end{itemize}

\subsection{Problem Definition} \label{sec:problem_definition}

The FL scenario comprises a set C of N clients where each client c $\in$ C possesses a dataset $D_c$. Before training starts, the server initializes a global model (usually a neural network) with parameters $\omega_0$ randomly generated, utilizing a baseline from a similar task or training with a public dataset whenever available.

Federated training involves executing $\tau$ rounds where, in each round $t \in {1, 2, ..., \tau}$, a subset of clients $S_t$ of size k (0 $< k \leq N$) is chosen to train the model. Each client c $\in S_t$ receives the model $\omega_t$ and conducts one epoch of training concurrently using its local dataset $D_c$.

The optimization method typically employed is Stochastic Gradient Descent (SGD).

\begin{equation}
    \label{eq:sgd}
    \omega_{t+1}^n \leftarrow \omega_{t} - \alpha \cdot \nabla L(\omega_{t}, D_c)
\end{equation}

Where $\alpha$ represents the learning rate and L denotes the loss function.

In FedAvg \cite{brendan2016communication}, the primary baseline of FL, several epochs of local training are conducted in each round.

Following each training round on the clients, each client $c$ transmits its updated parameters $\omega_{t+1}^c$ to the server. Subsequently, the server aggregates the received parameters into a new global model with parameters $\omega_{t+1}$ (using, for instance, the average of the received parameters).

The expectation is that the new model will synthesize all the knowledge acquired from each client’s training. However, meeting this expectation can prove challenging when the clients exhibit significant dissimilarity from one another.

\subsection{Client Selection} \label{sec:client_selection}

Client selection is a critical aspect of FL. Bandwidth limitations for exchanging information between clients and servers present a bottleneck that compromises the scalability of involving all clients in every round. \cite{cho2020client} demonstrated how a good client selection strategy can accelerate convergence by directing training to clients with the highest loss values.

This selection method begins by choosing a larger number of clients $k'$ ($k < k' \leq N$). Each of these k' clients receives the model parameters $\omega_{t}$ and evaluates the model's performance using its private dataset. Each client then returns its loss values to the server, which selects the k clients with the highest loss values to train in that round.

\subsection{Recover After Attack} \label{sec:recover}

Model Poisoning Attacks can be particularly destructive and permanently disrupt training progress. Our experiments have shown that, even after several successful rounds without attacks, training may fail to recover its accuracy following a severe attack.

To address this issue, FPP proposes canceling a round in the event of excessive deterioration in the model's performance. This involves saving the last approved checkpoint and reverting to it after a round that worsens metrics beyond a threshold, discarding the affected model.

To address this issue, FPP proposes discarding a round result and recovering a prior model checkpoint in the event of a deterioration in the model's performance. The model performance is estimated using the loss values received during the client selection, using the average of these loss values as an estimation of the global model's performance.

The main advantage of this approach lies in using loss values received from clients to evaluate the previous round, employing real data, and eliminating the need for a public dataset, which may be unavailable or too dissimilar from real data.

FPP defines a hyperparameter $\gamma$ as the rate of tolerated loss increment from which the training is considered irremediably prejudiced. So, if $e_t$ is the estimated loss of the global model at the round t, the model will be discarded if $e_{t} > e_{t-1} \cdot \gamma$ and the last approved checkpoint is recovered, preserving the training progress.

\subsection{Reputation Value}\label{sec:reputation}

To safeguard training by avoiding the selection of malicious clients, and mitigating the recurrence of checkpoint recovers, the FPP proposes a weight in the client selection method called reputation. This value is used as a weight during the first stage of customer selection. Thus, clients with higher reputations have a greater likelihood of being selected to evaluate the model.

Each client c initiates training with a maximum reputation value of $r_{c} = 1$ (an optimistic view where all clients are considered trustworthy), which is then updated as the client participates in rounds, either increasing or decreasing depending on round evaluations.

Upon detecting an adverse event ($e_{t} > e_{t-1} \cdot \gamma$), each client c that participated in the prior round t (c $\in S_t$) has its reputation value penalized according to $r_{c} \leftarrow r_{c} \cdot \delta_{p}$. Where $\delta_{p}$ represents the reputation penalty rate, a value below 1 (lower values indicate greater penalties).

When a round is deemed successful in the evaluation ($e_{t} \leq e_{t+1} \cdot \gamma$), the reputation values of clients who trained in round t $c \in S_t$ are restored as follows: $r_{i} \leftarrow min(1, r_{i} \cdot \delta_{r})$, where $\delta_{r}$ represents the reputation recovery rate (another hyperparameter, slightly higher than 1).

Throughout the rounds, as different subsets of clients train, honest clients participating in the same round as an attacker will momentarily have their reputation values reduced. However, these values will be restored as they participate in other successful rounds. Conversely, attackers will consistently be involved in problematic rounds, enduring increasingly severe penalties, making their selection for new rounds exceedingly rare.

\subsection{Secure Aggregation}\label{sec:secure_aggregation}

The FFP was designed to resist poisoning attacks without needing to access individual clients' gradients since it evaluates only the aggregated model. This makes the FFP fully compatible with Secure Aggregation, which mask the gradients sent by each client while preserving the aggregation average, increasing the privacy of the participant's data.

\begin{figure}[h]
\centering
    \includegraphics[width=0.35\textwidth]{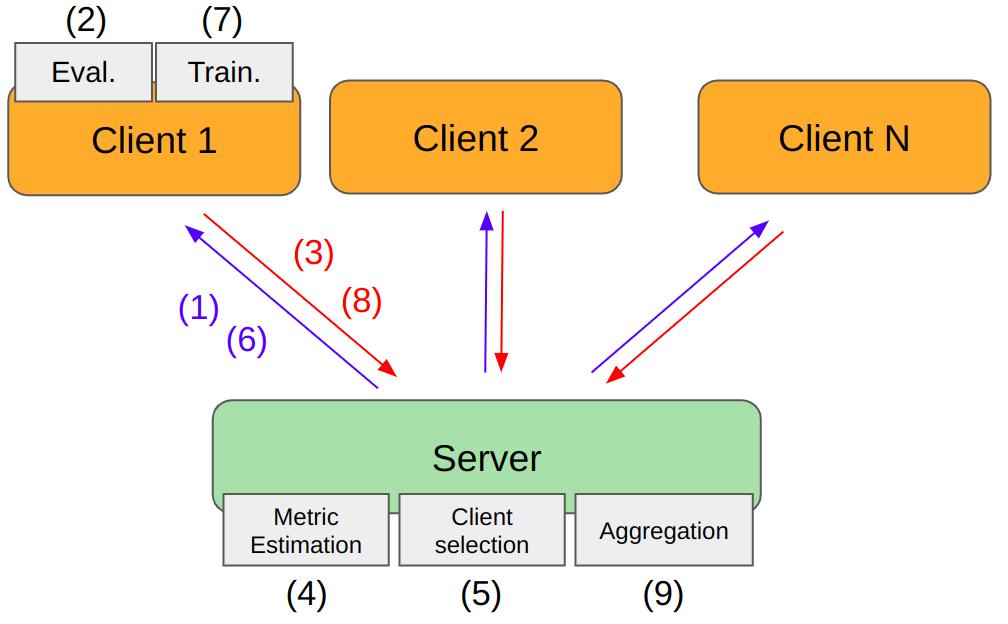}
    \caption{Process of a training round in the FPP.}
    \label{fig:diagrama}
\end{figure}

Figure \ref{fig:diagrama} illustrates the dynamic of a round in the FPP. The server sends the model parameters to selected clients (1), who evaluate the model using their local data (2) and return loss values to the server (3). With these values, the server estimates the global loss and updates participants' reputation values from the prior round (4). Subsequently, the server selects clients with the highest reputation values (5) and requests they conduct an epoch of training (6). After training completion (7), clients return the model gradients (8). The server aggregates the received gradients and applies the result to the global model, generating a new model (9).

\subsection{Algorithm}

Algorithm \ref{alg:principal} outlines the primary flow of FPP (section \ref{sec:problem_definition}). In each round, a subset of clients is selected to train in parallel, and the trained models are aggregated using the average on the server to create a new global model. Secure aggregation may be used to mask the weights $\omega_t^c$, generating masked weights $\omega_t^{*c}$. Secure aggregation ensures that $\sum_{c \in S_{t}}\omega_t^c = \sum_{c \in S_{t}}\omega_t^{*c}$.

The algorithm requires the following hyperparameters of the following table:

\begin{table}[h]
\centering
\begin{tabular}{|c|l|}
\hline
\textbf{Parameter} & \textbf{Definition}                                             \\ \hline
$\gamma$           & Damage threshold for attack detection ($\gamma > 1$).                 \\ \hline
$\delta_p$         & Reputation penalty in case of attack ($\delta_p < 1$).          \\ \hline
$\delta_r$         & Reputation recover after a successful round ($\delta_P > 1$).                        \\ \hline
k                  & Number of training clients in each round.                       \\ \hline
k'                 & Number of clients selected in the initial selection. \\ \hline
\end{tabular}
\end{table}

\begin{algorithm}
\caption{Main Flow of FPP}\label{alg:principal}
\textbf{FederatedTraining($\gamma, \delta_p, \delta_r, k, k'$):}
\begin{algorithmic}
\State $\omega_0 \leftarrow RandomInitialization()$
\State $C = \{c_1, c_2, ..., c_N\}$  \Comment{Clients set}
\State $R = \{r_i = 1 \vert 1 \leq i \leq N\}$  \Comment{Reputation values}
\State $e^{*} \leftarrow None$  \Comment{Last approved model's loss}
\State $\omega^{*} \leftarrow None$  \Comment{Last approved model's parameters}
\While{t in 1, 2, 3,..., $\tau$}
    \State $S^t \leftarrow ClientSelection$  \Comment{See Algorithm \ref{alg:selecao}}
    \While{c in $S_t$}  \Comment{Each client in parallel}
        \State $\omega_t^{*c} \leftarrow LocalTraining(client=c, model=\omega_{t-1})$
    \EndWhile
    \State $\omega_{t+1} \leftarrow \frac{1}{k} \sum_{c \in S_{t}} \omega_{t}^{*c}$  \Comment{Aggregation}
\EndWhile
\end{algorithmic}
\end{algorithm}

\begin{algorithm}
\caption{Client Selection Algorithm}\label{alg:selecao}
\textbf{ClientSelection}
\begin{algorithmic}
\State $S_t' \leftarrow$ SubSet(clients=C, size=k', weight=R)
\While{c in $S_t'$}  \Comment{In each client in parallel}
    \State $loss_c \leftarrow$ Evaluates($\omega_{t-1}$)
\EndWhile
\If{First evaluation in the current round}
    \State $e_t \leftarrow$ Average(\{$loss_c \vert c \in S_t'$\})  \Comment{Model's metric estimation in the round t}
    \State Evaluate\&Recover($e_t$)  \Comment{See Algorithm \ref{alg:avaliacao}}
    \If{Previous model was discarded}
        \State Restart Client Selection
    \EndIf
\EndIf
\State \textbf{return} the k clients with the highest loss values
\end{algorithmic}
\end{algorithm}

Algorithm \ref{alg:selecao} provides details on how client selection is conducted (section \ref{sec:client_selection}). The process begins by selecting k' clients to evaluate the model's performance. The server utilizes these evaluations to assess the previous round and determine if an attack occurred. If the model is approved, the k clients with the highest loss values are returned to Algorithm \ref{alg:principal}. If the model's metric deteriorates beyond $\gamma$, the model is discarded, the last approved model is restored, and the client selection process restarts without needing to reevaluate the model (since it was previously approved).

\begin{algorithm}[h]
\caption{Evaluation and Model Recovering Algorithm}\label{alg:avaliacao}
\textbf{Evaluate\&Recover($e_t$)}
\begin{algorithmic}
\If{$t>0$ e $e_{t} > e^{*} \cdot \gamma$}   \Comment{Metric deteriorated above the tolerated threshold}
    \State $\omega_t \leftarrow \omega^{*}$   \Comment{Recover the last approved checkpoint}
    \While{c in $S_{t-1}$}:  \Comment{For each client participant in the last round}
        \State $r_c \leftarrow r_c \cdot \delta_p$   \Comment{Apply the severe penalty}
    \EndWhile
\ElsIf{$t=0$ or $e_{t} \leq e^{*} \cdot \gamma$}  \Comment{Checkpoint approved}
    \State $\omega^{*} \leftarrow w_{t-1}$   \Comment{Save the checkpoint}
    \State $e^{*} \leftarrow e_{t-1}$   \Comment{Update the saved checkpoint metric}
    \While{c in $S_{t-1}$}:   \Comment{For each client participant in the last round}
        \State $r_c \leftarrow r_c \cdot \delta_r$  \Comment{Increase the reputation value}
    \EndWhile
\EndIf
\end{algorithmic}
\end{algorithm}

Algorithm \ref{alg:avaliacao} outlines how models are evaluated in each round (section \ref{sec:recover} and \ref{sec:reputation}). This function receives the estimation of the global model's loss and applies a threshold to detect attacks, recovering the model if necessary. Additionally, this algorithm updates the reputation values of clients who participated in the previous round ($S_{t-1}$).

\section{Experiment Setup} \label{sec:setup}

This section presents the setup of the experiments to validate FPP and compare it with FedAvg \cite{brendan2016communication} (the FL baseline), PoC \cite{cho2020client} (an approach with rapid convergence), and two aggregation strategies resilient to attacks by \cite{yin2018byzantine}: Trimmed Mean and Median. These five approaches were utilized to train a model using FL, both with and without attackers among the honest clients.

The experiments were conducted to verify the behavior of the approaches in the presence of malicious participants, particularly performing model poisoning (corrupting the gradients returned to the server).

\subsection{Dockerized Implementation}

The implementation proposed to perform the experiments consisted of a Docker environment where each client is isolated in a Docker Container and communicates to a server in a FL training. Each container stores the dataset and the communication between client and server is done using the HTTP protocol. The implementation is available on GitHub\footnote{github.com/NRiccieri/FastPrivateProtected}.

Due to hardware limitations, we run the experiments using 12 clients, each in a separate Docker container controlled by a Docker Swarm.

\subsection{Dataset and Model Architecture}

The dataset was generated using LEAF \cite{caldas2018leaf}, a framework for FL designed to generate datasets with appropriate properties for validating and comparing FPP with other approaches. The synthetic dataset comprises 10 numeric features and a 6-class categorical output.

The chosen model architecture was a multilayer perceptron (neural network) with 3 hidden layers of 256 rectified linear unit (ReLU) neurons and a softmax output layer with 6 neurons. Stochastic Gradient Descent (SGD) was employed as the optimizer, with a learning rate of $10^{-3}$.

All experiments start from the same checkpoint (architecture and parameters), thereby eliminating one possible source of variation in the analysis.

\subsection{Experiments}

The initial test evaluated the performance of FPP considering only honest clients. In this scenario, without attackers, it was anticipated that all rounds would be approved, and no client would incur penalties for their reputation values.

The first simulated model poisoning attack involved a malicious participant sending noise (following a normal distribution) instead of real gradients.
 
Finally, we tested the approach against a malicious client training the model but returning the gradient in the opposed direction, maximizing the loss instead of minimizing it.

The hyperparameters we used were determined using grid search and are as follows: k=6, k'=9, $\gamma = 1.25, \delta_p = 0.85$ and $\delta_r = 1.2$.

\section{Results and Analysis}

\begin{figure}[ht]
\centering
    \includegraphics[width=0.45\textwidth]{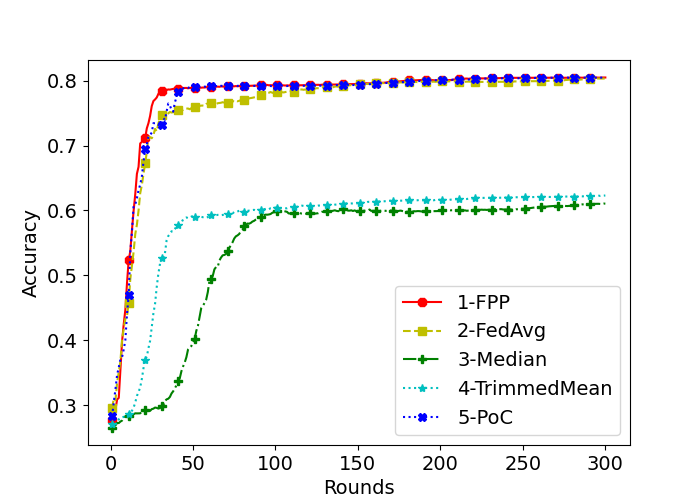}
    \caption{Accuracy per round in the training with only honest clients.}
    \label{fig:base}
\end{figure}

In the absence of attackers (Figure \ref{fig:base}), the FPP, PoC, and FedAvg approaches exhibit similar behavior, stabilizing at an accuracy value of 80\%, with PoC and FPP being slightly faster than FedAvg. Other strategies that resist attacks have lower accuracy, around 60\%.

\begin{figure}[ht]
\centering
    \includegraphics[width=0.45\textwidth]{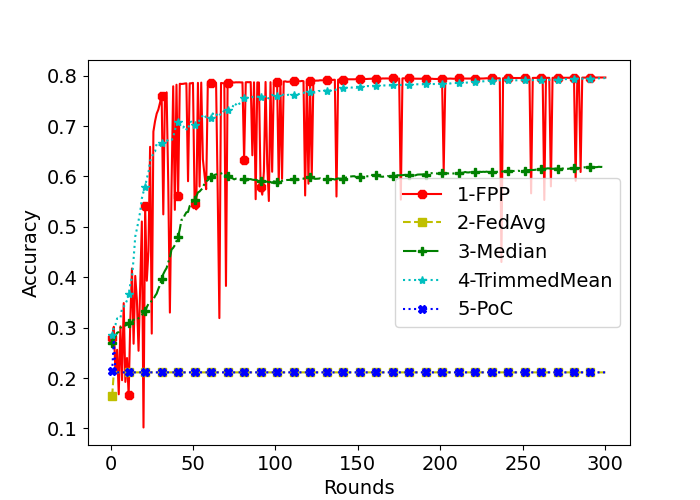}
    \caption{Accuracy per round in the training with one attacker sending noise.}
    \label{fig:noise}
\end{figure}

Figure \ref{fig:noise} illustrates the outcome when an attacker (among 11 honest clients) conducts a model poisoning attack by sending random noise instead of legitimate gradients. The graph shows that only FPP and TrimmedMean converge to an accuracy near 80\%. TrimmedMean achieves this by eliminating extreme gradients, while FPP recovers after attacks without evaluating or sorting gradients, enabling Secure Aggregation to protect participants' data. The Median strategy remains attack-resistant but only reaches about 60\% accuracy, whereas the attacks completely disrupt PoC and FedAvg.

\begin{figure}[ht]
\centering
    \includegraphics[width=0.45\textwidth]{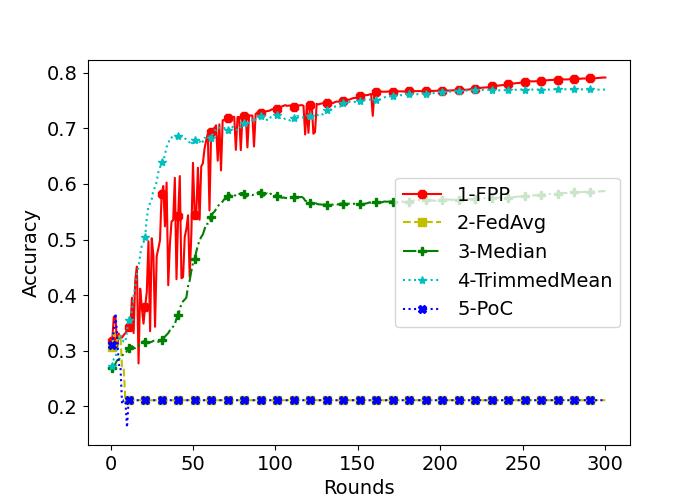}
    \caption{Accuracy per round in the training with one attacker maximizing the loss.}
    \label{fig:scale}
\end{figure}

Similar results can be seen in the case of attacks where the malicious participant trains the model but corrupts the gradients sent by reversing the direction to maximize loss instead of minimizing it (Figure \ref{fig:scale}). FPP manages to recover from the attacks by following the TrimmedMean strategy, which removes outlier values.

It is also noted in both attack scenarios how FPP exhibits much less noise (indicating fewer occurrences of attacks) as the training progresses, indicating the success of the strategy to pre-selected clients based on reputation value. In these experiments, the reputation values of the attacker remained around 0.05, while the reputation values of honest clients were greater than or equal to 0.85 (penalty rate in case of attack).

As we conducted these experiments with only 12 clients due to hardware limitations, the experiment with only honest clients was unable to demonstrate the true advantage of client selection biased by the highest loss value in convergence speed, given that a significant portion of clients participated in training in each round.

To demonstrate this advantage, we conducted an additional experiment running the algorithm in a local environment with all clients running in the same process with 345 clients. A new dataset was created using LEAF for this scenario, also with 10 input features and 6 output classes. In each round, only 7 clients were selected for training, with 32 clients pre-selected in both FPP and PoC.

\begin{figure}[ht]
\centering
    \includegraphics[width=0.45\textwidth]{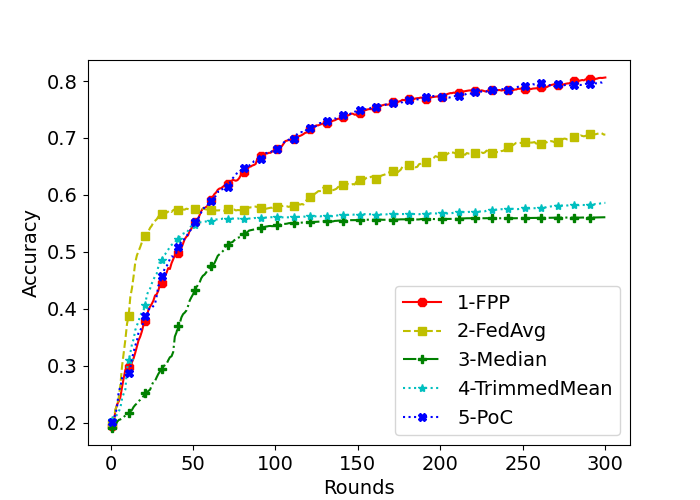}
    \caption{Accuracy per round in the training with 345 honest clients.}
    \label{fig:base345}
\end{figure}

The results of this experiment are shown in Figure \ref{fig:base345}, where it can be observed how FPP and PoC converge faster than FedAvg. This occurs because client selection needs to be more stringent when only a small portion of clients train in each round (in this experiment, only 2\% of the total, i.e., 7 out of 345).

\section{Conclusion} \label{sec:conclusao}

In this paper, we introduced FPP, a novel approach for FL designed to achieve rapid training convergence even with non-iid data, while also being resilient to model poisoning attacks and enabling the use of secure aggregation to preserve data privacy.

Our experiments demonstrated that FPP exhibits consistent training progress, converging in the presence of attackers attempting to disrupt the training by corrupting their gradients.

Furthermore, FPP utilizes a reputation value to mitigate the probability of selecting a malicious client for training, effectively distinguishing them from honest clients without compromising the privacy of individual gradients.

As future work, we propose to evaluate using other model architectures and tasks such as CNN or Language Models. We also propose to validate the protection against more sophisticated attacks, such as backdoor attacks \cite{bagdasaryan2020backdoor}, or strategies aimed at compromising privacy even with secure aggregation \cite{kariyappa2023cocktail}.

\section{Acknowledgment}

This project was supported by the Ministry of Science, Technology, and Innovation of Brazil , with resources granted by the Federal Law 8.248 of October 23, 1991, under the PPI-Softex. The project was coordinated by Softex and published as Intelligent agents for mobile platforms based on Cognitive Architecture technology [01245.003479/2024-10].

%\begin{thebibliography}{00}
\bibliographystyle{plain}
\bibliography{conference}

\begin{thebibliography}{10}

\bibitem{bagdasaryan2020backdoor}
Eugene Bagdasaryan, Andreas Veit, Yiqing Hua, Deborah Estrin, and Vitaly Shmatikov.
\newblock How to backdoor federated learning.
\newblock In {\em International conference on artificial intelligence and statistics}, pages 2938--2948. PMLR, 2020.

\bibitem{bonawitz2017practical}
Keith Bonawitz, Vladimir Ivanov, Ben Kreuter, Antonio Marcedone, H~Brendan McMahan, Sarvar Patel, Daniel Ramage, Aaron Segal, and Karn Seth.
\newblock Practical secure aggregation for privacy-preserving machine learning.
\newblock In {\em proceedings of the 2017 ACM SIGSAC Conference on Computer and Communications Security}, pages 1175--1191, 2017.

\bibitem{brendan2016communication}
H~Brendan~McMahan, Eider Moore, Daniel Ramage, Seth Hampson, and Blaise Ag{\"u}era~y Arcas.
\newblock Communication-efficient learning of deep networks from decentralized data.
\newblock {\em arXiv e-prints}, pages arXiv--1602, 2016.

\bibitem{caldas2018leaf}
Sebastian Caldas, Sai Meher~Karthik Duddu, Peter Wu, Tian Li, Jakub Kone{\v{c}}n{\`y}, H~Brendan McMahan, Virginia Smith, and Ameet Talwalkar.
\newblock Leaf: A benchmark for federated settings.
\newblock {\em arXiv preprint arXiv:1812.01097}, 2018.

\bibitem{cho2020client}
Yae~Jee Cho, Jianyu Wang, and Gauri Joshi.
\newblock Client selection in federated learning: Convergence analysis and power-of-choice selection strategies.
\newblock {\em arXiv preprint arXiv:2010.01243}, 2020.

\bibitem{jiang2024research}
Xiaoyu Jiang, Ruichun Gu, and Huan Zhan.
\newblock Research on incentive mechanisms for anti-heterogeneous federated learning based on reputation and contribution.
\newblock {\em Electronic Research Archive}, 32(3):1731--1748, 2024.

\bibitem{kang2019incentive}
Jiawen Kang, Zehui Xiong, Dusit Niyato, Shengli Xie, and Junshan Zhang.
\newblock Incentive mechanism for reliable federated learning: A joint optimization approach to combining reputation and contract theory.
\newblock {\em IEEE Internet of Things Journal}, 6(6):10700--10714, 2019.

\bibitem{kariyappa2023cocktail}
Sanjay Kariyappa, Chuan Guo, Kiwan Maeng, Wenjie Xiong, G~Edward Suh, Moinuddin~K Qureshi, and Hsien-Hsin~S Lee.
\newblock Cocktail party attack: Breaking aggregation-based privacy in federated learning using independent component analysis.
\newblock In {\em International Conference on Machine Learning}, pages 15884--15899. PMLR, 2023.

\bibitem{lyu2020threats}
Lingjuan Lyu, Han Yu, and Qiang Yang.
\newblock Threats to federated learning: A survey.
\newblock {\em arXiv preprint arXiv:2003.02133}, 2020.

\bibitem{nair2023robust}
Akarsh~K Nair, Ebin~Deni Raj, and Jayakrushna Sahoo.
\newblock A robust analysis of adversarial attacks on federated learning environments.
\newblock {\em Computer Standards \& Interfaces}, page 103723, 2023.

\bibitem{pillutla2019robust}
Krishna Pillutla, Sham~M Kakade, and Zaid Harchaoui.
\newblock Robust aggregation for federated learning.
\newblock {\em arXiv preprint arXiv:1912.13445}, 2019.

\bibitem{wang2020optimizing}
Hao Wang, Zakhary Kaplan, Di~Niu, and Baochun Li.
\newblock Optimizing federated learning on non-iid data with reinforcement learning.
\newblock In {\em IEEE INFOCOM 2020-IEEE Conference on Computer Communications}, pages 1698--1707. IEEE, 2020.

\bibitem{xie2018generalized}
Cong Xie, Oluwasanmi Koyejo, and Indranil Gupta.
\newblock Generalized byzantine-tolerant sgd.
\newblock {\em arXiv preprint arXiv:1802.10116}, 2018.

\bibitem{yin2018byzantine}
Dong Yin, Yudong Chen, Ramchandran Kannan, and Peter Bartlett.
\newblock Byzantine-robust distributed learning: Towards optimal statistical rates.
\newblock In {\em International Conference on Machine Learning}, pages 5650--5659. PMLR, 2018.

\bibitem{zhang2021client}
Wenyu Zhang, Xiumin Wang, Pan Zhou, Weiwei Wu, and Xinglin Zhang.
\newblock Client selection for federated learning with non-iid data in mobile edge computing.
\newblock {\em IEEE Access}, 9:24462--24474, 2021.

\bibitem{zhang2023reputation}
Zhibo Zhang, Pengfei Li, Ahmed Y~Al Hammadi, Fusen Guo, Ernesto Damiani, and Chan~Yeob Yeun.
\newblock Reputation-based federated learning defense to mitigate threats in eeg signal classification.
\newblock {\em arXiv preprint arXiv:2401.01896}, 2023.

\bibitem{zhu2019deep}
Ligeng Zhu, Zhijian Liu, and Song Han.
\newblock Deep leakage from gradients.
\newblock {\em Advances in neural information processing systems}, 32, 2019.

\end{thebibliography}
%\end{thebibliography}

\end{document}